\documentclass[lettersize,journal]{IEEEtran}
\usepackage{amsmath,amsfonts}
\usepackage{algorithmic}
\usepackage{algorithm}
\usepackage{array}
\usepackage[caption=false,font=normalsize,labelfont=sf,textfont=sf]{subfig}
\usepackage{textcomp}
\usepackage{stfloats}
\usepackage{url}
\usepackage{verbatim}
\usepackage{graphicx}
\usepackage{cite}
\usepackage{amsmath}
\usepackage{amssymb}
\usepackage{mathtools}
\usepackage{amsthm}

\usepackage{microtype}
\usepackage{graphicx}
\usepackage{subcaption}
\usepackage{booktabs} % for professional tables

\usepackage[capitalize,noabbrev]{cleveref}

\usepackage{multirow}
\usepackage{makecell}
\usepackage{bm}

\theoremstyle{plain}
\newtheorem{theorem}{Theorem}[section]

\theoremstyle{definition}

\theoremstyle{remark}

\begin{document}

\title{  Bipartite Graph Attention-based Clustering \\ for Large-scale scRNA-seq Data}

\author{Zhuomin Liang, Liang Bai, Xian Yang
        % <-this % stops a space
\IEEEcompsocitemizethanks{\IEEEcompsocthanksitem
Z. Liang, L. Bai and J. Liang are with Institute of Intelligent Information Processing, Shanxi University, Taiyuan, 030006, China (Corresponding author: Liang Bai)\protect\\
Email: sxzhuominliang@163.com, bailiang@sxu.edu.cn, ljy@sxu.edu.cn

X. Yang is  with Alliance Manchester Business School, The University of Manchester, Manchester, M13 9PL, UK\protect\\
Email: xian.yang@manchester.ac.uk
}
\thanks{}% <-this % stops a space
\thanks{}

}

\markboth{Journal of \LaTeX\ Class Files,~Vol.~14, No.~8, August~2021}%
{Shell \MakeLowercase{\textit{et al.}}: A Sample Article Using IEEEtran.cls for IEEE Journals}

\IEEEpubid{0000--0000/00\$00.00~\copyright~2021 IEEE}

\maketitle

\begin{abstract}
scRNA-seq clustering is a critical task for analyzing single-cell RNA sequencing (scRNA-seq) data, as it groups cells with similar gene expression profiles. Transformers, as powerful foundational models, have been applied to scRNA-seq clustering. Their self-attention mechanism automatically assigns higher attention weights to cells within the same cluster, enhancing the distinction between clusters.
% However, scRNA-seq data lack the sequential structure required by Transformers. 
Existing methods for scRNA-seq clustering, such as graph transformer-based models, treat each cell as a token in a sequence. Their computational and space complexities are $\mathcal{O}(n^2)$ with respect to the number of cells, limiting their applicability to large-scale scRNA-seq datasets.
To address this challenge, we propose a Bipartite Graph Transformer-based clustering model (BGFormer) for scRNA-seq data. We introduce a set of learnable anchor tokens as shared reference points to represent the entire dataset. A bipartite graph attention mechanism is introduced to learn the similarity between cells and anchor tokens, bringing cells of the same class closer together in the embedding space. BGFormer achieves linear computational complexity with respect to the number of cells, making it scalable to large datasets. 
Experimental results on multiple large-scale scRNA-seq datasets demonstrate the effectiveness and scalability of BGFormer.
\end{abstract}


\section{Introduction}
\IEEEPARstart{s}cRNA-seq clustering is the a fundamental step in single-cell RNA sequencing (scRNA-seq) data analysis. By grouping cells with similar transcriptional profiles, researchers can identify distinct cell types and subpopulations \cite{scrnadata}, and explore cellular heterogeneity \cite{massively}.
% Single-cell RNA sequencing (scRNA-seq) data provides a unique opportunity to characterize cellular complexity and heterogeneity \cite{scrnadata}. Through the analysis of scRNA-seq data, researchers can effectively deconvolute the complexity of biological systems \cite{complex}, precisely delineate diverse cell types \cite{newcelltype}, and reconstruct developmental trajectories \cite{apply}. Clustering analysis is a fundamental step in scRNA-seq data analysis, grouping cells with similar transcriptional profiles. 
However, the high dimensionality, sparsity, and technical variability inherent in scRNA-seq data pose significant challenges to existing clustering methods.

\begin{figure}[!t]
    \centering
    \includegraphics[width=\columnwidth]{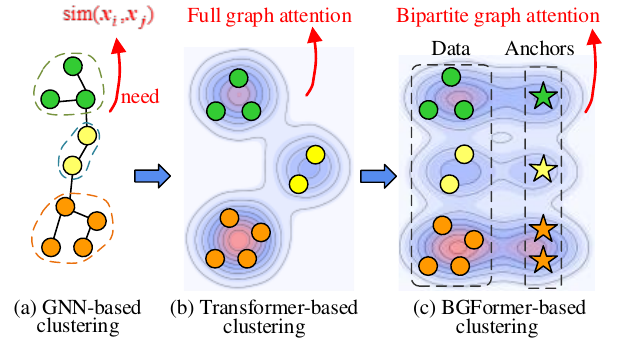}
    \caption{Comparison of different clustering methods.}
    \label{fig:att_framework}
      \vskip -0.1in
\end{figure}

% Graph-based clustering methods are effective to address these challenges by representing cells as nodes in a graph, where edges represent pairwise similarities \cite{scgnn}.

Graph-based clustering methods provide an effective solution to these challenges by leveraging cell relationships to group similar cells into clusters \cite{scgnn}. Both Graph Neural Networks (GNNs) and Transformers, which are powerful tools for modeling relationships, have been employed for scRNA-seq clustering. GNN-based methods rely on explicit relationships to guide the encoding of cell information \cite{scgnn, scG_cluster}. As illustrated in Fig.~\ref{fig:att_framework}(a), these methods require calculating the similarity among cells to construct the k-nearest neighbor (kNN) graph, as the relationships between cells are not inherently available.
% Since the relationships among cells are not directly available, these methods typically construct a k-nearest neighbor (kNN) graph based on cell similarity. 
However, the high dimensionality and sparsity of scRNA-seq data often undermine the effectiveness of similarity metrics, limiting the performance of these clustering methods. In contrast, Transformer-based methods can better handle these challenges due to their ability to capture complex relationships without relying on traditional similarity metrics. These methods model the input as a sequence of tokens, with each cell treated as an individual token \cite{transformers_cell1, scgraphformer}. As shown in Fig. \ref{fig:att_framework}(b), the self-attention mechanism learns a full graph attention to implicitly capture the similarities among cells. By adaptively adjusting the similarity matrix, the differences between clusters are increased, making it easier to identify cell groups without explicit guidance.
% the self-attention mechanism automatically learns the similarities among cells, identifying groups of similar cells without explicit guidance. 
Although these methods eliminate the need for kNN graph construction, the quadratic computational complexity of self-attention with respect to the number of cells makes them prohibitively expensive for large-scale datasets.
\IEEEpubidadjcol

We propose  a novel \textbf{B}ipartite \textbf{G}raph Trans\textbf{Former}-based clustering model (BGFormer) for large-scale scRNA-seq data. The core of BGFormer is to shift the similarity calculation for clustering from between cells to between cells and learnable anchor tokens, as illustrated in Fig. \ref{fig:att_framework}(c). These anchors serve as reference points for the entire dataset, allowing similar cells to aggregate similar global information. Since its number is much smaller than the number of cells, the computational complexity is reduced from $\mathcal{O}(n^2)$ to $\mathcal{O}(n)$, where $n$ is the number of cells.  The anchor tokens are shared across mini-batches, making BGFormer inherently suitable for training on large-scale scRNA-seq datasets. The main contributions of this paper are as follows:
% reformulate the fully connected self-attention mechanism into a bipartite graph structure, thereby alleviating the quadratic computational and memory costs of standard Transformers. Specifically, scBGFormer introduce a set of learnable anchor tokens that serve as latent representatives of the all cells. Instead of explicitly modeling all pairwise cell–cell interactions, cell embeddings are learned by aggregating information from these anchor tokens. 
% % To ensure that the anchor tokens effectively capture global cellular information, we design an anchor token learning module with a reconstruction-based objective, which optimizes the anchors to reconstruct the original cell expression profiles. Based on the learned anchor tokens, we further replace the conventional self-attention in Graph Transformers with a bipartite graph attention mechanism, where interactions are restricted to cell–anchor pairs. 
% This design reduces the computational complexity from quadratic to linear with respect to the number of cells. 

\begin{itemize}
    \item We introduce an efficient clustering method for large-scale scRNA-seq data, achieving linear computational complexity with respect to the number of cells.

    \item We introduce a bipartite graph attention mechanism that learns similarity for clustering from the relationships between cells and anchor tokens.
    
    \item Experiments on large-scale scRNA-seq datasets demonstrate that scBGFormer achieves lower computational cost and superior clustering performance.
\end{itemize}

% In this way, its computational cost increases linearly with the number of cells, which can be easily applied to large-scale datasets. To learn effective anchor tokens, reconstruction objective is set to reconstruct all cells. With these anchors, we replace the self-attention in graph Transformer with the bipartite graph attention. 

\section{Related Works}
% Clustering analysis is a fundamental step in scRNA-seq data analysis \cite{scice}, aiming to group cells with similar gene expression profiles.  
\subsection{Graph-Based scRNA-seq Clustering}
Graph-based clustering methods are widely used for scRNA-seq analysis, as they model relationships among cells. Since such relationships are not directly available, these methods rely on similarity metrics to construct explicit graph structures, such as those used in k-means \cite{kmeans} algorithms. Methods like Louvain and Leiden \cite{louvain} are then applied to identify cell clusters. scICE \cite{scice} further enhance the reliability of clustering by evaluating multi-cluster label consistency. With the development of graph neural networks (GNNs) \cite{gcn}, many methods group cells by bringing similar cells closer in the embedding space. Methods including scGAE \cite{scgae}, scGCL \cite{scgcl}, scTAG \cite{zinb}, and CCST \cite{ccst} adopt autoencoder or contrastive learning frameworks to learn cell embeddings in an unsupervised manner. scG-cluster \cite{scG_cluster} utilizes multiple graph convolution kernels, while scGNN \cite{scgnn} and scSimGCL \cite{scSimGCL} iteratively refine the cell graph to reduce noise. scTPF \cite{scTPF} explores the interaction between local and global latent configurations. 
% These methods rely on similarity metrics to construct explicit graph structures.
However, due to the low-quality of scRNA-seq data, similarity metrics often fail to accurately reflect cell groupings, limiting clustering performance. Transformer-based methods, such as scGraphformer \cite{scgraphformer}, TOSICA \cite{transformers_cell2} and single-cell transformer \cite{transformers_cell1} treat cells as tokens in a sequence and employ a Transformer/Graph Transformer network to capture interactions among cells through the self-attention mechanism. It learns the relationships between any cell pairs, resulting in high computational costs and limited scalability to large-scale scRNA-seq datasets.

\subsection{Large-scale scRNA-seq Clustering}
Mini-batch processing is an effective strategy for handling large-scale scRNA-seq data. Deep embedding–based methods, which encode each cell independently, are naturally suitable for this strategy. For example, scDCC \cite{scDCC}, scMDC \cite{scMDC}, IDEC \cite{idec}, and scVAE \cite{scvae} encode each cell using fully connected neural networks. Due to the intrinsic noise and sparsity of scRNA-seq data, such independent encoding fails to capture inter-class differences. MetaQ \cite{metaq} improves scalability by clustering metacells, but overlooks fine-grained cellular heterogeneity. To reduce the complexity of Transformers, several method employ low-rank approximations \cite{linformer, nodeformer}, allowing them to efficiently process long sequences. However, in large-scale scRNA-seq clustering tasks, the sequence length can become excessively large, limiting the effectiveness of these methods.
To alleviate this issue, some methods also adopt mini-batch training strategy and construct graphs within each batch \cite{scSimGCL, scgraphformer}. However, this strategy neglects relationships beyond the batch, resulting in the loss of global structural information, which degrades clustering performance. 
% MetaQ \cite{metaq} addresses scalability by compressing the dataset into a set of representative metacells and performing clustering at the metacell level. While computationally efficient, this strategy inevitably overlooks fine-grained cellular heterogeneity.  
To mitigate this limitation, we design a BGFormer-based clustering model, where shared global tokens enable efficient modeling of long-range interactions across all cells with linear complexity.

\section{Preliminaries}
\subsection{Notations}
Given a scRNA-seq dataset comprising $n$ cells and $d'$ genes, we represent the raw data as a gene expression matrix $\hat{\bm{X}} \in \mathbb{R}^{n \times d'}$, where $x_{i,j}$ denotes the expression count of gene $j$ in cell $i$. Due to the high dropout rate in gene expression profiles, data filtering and quality control are performed to retain highly variable genes. After log transformation and normalization, the processed data are represented as ${\bm{X}} \in \mathbb{R}^{n \times d}$ for downstream tasks, where $d$ denotes the number of selected top-ranked genes.

% \subsection{Self-Attention in Transformers}
\subsection{Transformer-based scRNA-seq Clustering}
\begin{figure}
    \centering
    \includegraphics[width=0.95\columnwidth]{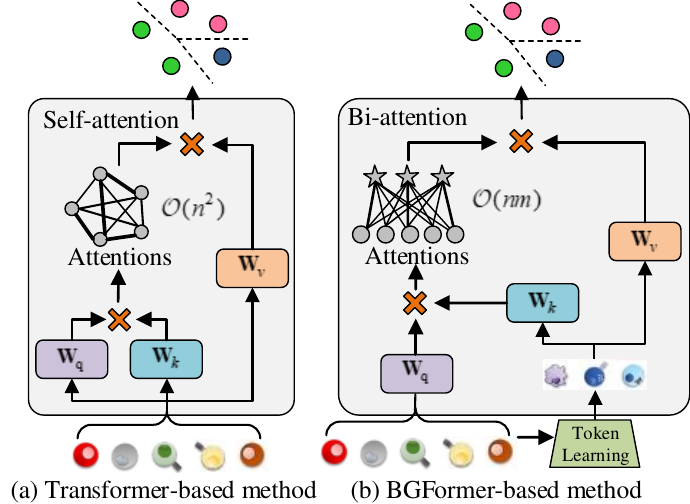}
    \caption{Comparison of the attention mechanism in traditional Transformer and BGFormer-based clustering model.}
    \label{fig:atten_c}
     \vskip -0.1in
\end{figure}

The goal of scRNA-seq clustering is to ensure that similar cells have the same predictions in an unsupervised manner. Graph-based clustering methods are formulated as:
\begin{equation}
    \hat{\bm{Y}} = f(\bm{A}\bm{X}),
\end{equation}
where $\bm{A}$ denotes the similarity matrix among cells, and $\hat{\bm{Y}}$ represents the predicted labels. As widely used backbone architecture, Transformers have been employed to implement this process. As shown in Fig. \ref{fig:atten_c}(a), the self-attention mechanism in Transformers automatically learns the similarity matrix as:
\begin{equation}
    \bm{A} = softmax (\frac{\bm{Q} \bm{K}^T}{\sqrt{d_k}}),
\label{eq:emb_weight}
\end{equation}
where $d_k$ is the dimensionality of $\bm{Q}$, $\bm{Q} = \bm{X} \bm{W}_Q$ and $\bm{K} = \bm{X} \bm{W}_K$. Here, $\bm{W}_Q$ and $\bm{W}_K$ are learnable projection matrices that map cell information into a low-dimensional space, reducing the impact of low-quality data and allowing the model to adaptively learn the similarity matrix during training. $\bm{X}$ represents the sequence of all cells, with each cell treated as an individual token. The matrix $\bm{A}$ computed in Eq. (\ref{eq:emb_weight}) can be interpreted as a fully connected graph, where cell pairs with higher similarity are assigned greater weights. Cell information is then propagated over the matrix $\bm{A}$, which is formulated by:
\begin{equation}
    \hat{\bm{Z}} = \bm{A} \bm{V}, \bm{V} = \bm{X} \bm{W}_V,
\label{eq:emb}
\end{equation}
where $\bm{W}_V$ is learnable parameters, $\hat{\bm{Z}}$ represents the learned cell embeddings. 
% This process enhances the similarity between cells, facilitating more accurate identification of potential groupings within the data. 
To perform clustering, existing methods typically introduce a set of learnable cluster centroids $\{\bm{\mu}_j\}_{j=1}^K$ and adopt the Deep Embedded Clustering (DEC) objective \cite{DEC}. The DEC loss is defined as the Kullback–Leibler (KL) divergence between an auxiliary target distribution and the soft assignment distribution, which is then formulated as:  
\begin{equation}
    \mathcal{L}_{c} = KL(\bm{P}||\bm{Q})=\sum_i\sum_j p_{ij}log\frac{p_{ij}}{q_{ij}},
\label{eq:cluser}
\end{equation}  
where $p_{ij}$ is the soft assignment probability of cell $i$ and cluster centroid $j$, and $q_{ij}$ is the corresponding target distribution. After training, the final cluster label for each cell is obtained by selecting the centroid with the highest assignment probability.

While the above method is effective for scRNA-seq clustering, it becomes impractical for large-scale datasets. The self-attention mechanism involves all-to-all interactions, resulting in both computational and memory complexities of $\mathcal{O}(n^2)$. As the number of cells $n$ grows, the costs of training and inference become increasingly prohibitive. This motivates the development of more efficient attention mechanisms to handle such large datasets.

\section{Methods}

% \begin{figure}[!t]
%     \centering
%     \includegraphics[width=\columnwidth]{figs/attention_clusshu.pdf}
%     \caption{Illustrate of attention.}
%     \label{fig:att_framework}
% \end{figure}

% which reformulates self-attention from a fully connected graph over input tokens to a bipartite graph.

We propose a \textbf{B}ipartite \textbf{G}raph Trans\textbf{Former}-based clustering model (BGFormer), which calculates the similarity for clustering using a bipartite graph structure. 
As shown in Fig. \ref{fig:atten_c}(b), BGFormer introduces a set of learnable anchor tokens $\bm{U} = [\bm{u}_1, \bm{u}_2, \dots, \bm{u}_m]$, where $\bm{u}_i$ denotes the $i$-th anchor token and $m$ is the total number of anchors. These anchor tokens serve as reference points to construct the bipartite graph between cells and anchors, enabling computational costs that scale linearly with the number of cells. As a result, BGFormer can be efficiently applied to large-scale scRNA-seq data. However, the design of BGFormer poses two key challenges:
\begin{itemize}
    \item How can we effectively learn anchor tokens that provide meaningful global information for clustering?
     \item How can we design a scalable and efficient bipartite graph attention mechanism for clustering?
    % \item How to transform the fully connected self-attention graph into a bipartite graph?
    % \item How to design a bipartite graph Transformer?
\end{itemize}

% Based on this input, scEAFormer compresses the original $N$ cells into $M$ representative meta-cells, where $M \ll N$. By computing attention between cells and meta-cells, 
% % instead of performing full self-attention over all cell pairs, 
% the model refines cell embeddings with lower computational and space complexity. 

\begin{figure*}
    \centering
    \includegraphics[width=0.8\textwidth]{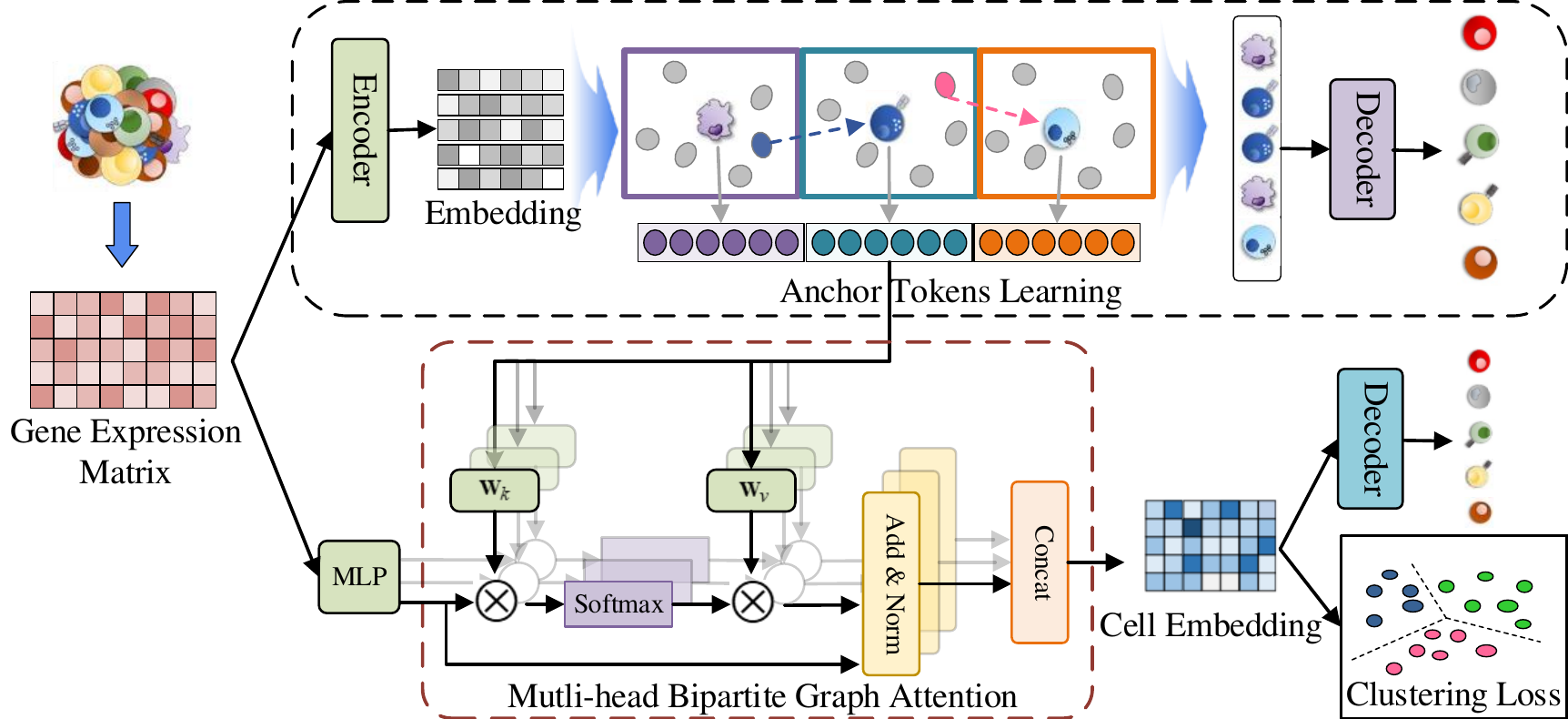}
    \caption{The framework of BGFormer-based clustering model. 
    % The codebook learning module compresses single cells into meta-cells, while the external attention module captures attention between single cells and meta-cells.
    }
    \label{fig:framework}
\end{figure*}

To address these challenges, the framework of BGFormer is designed as shown Fig. \ref{fig:framework}, consisting of two key modules, each corresponding to one of the challenges: the anchor token learning and the bipartite graph attention.

\subsection{Anchor Token Learning}
% Transformer architectures generally compute full self-attention across all cell pairs. This leads to a quadratic increase in computational complexity as the number of cells grows, making them impractical for large-scale RNA-seq data. 

% The anchor token learning module aims to learn a compact set of latent anchor tokens that can effectively represent all cells. By interacting with these anchors, each cell captures its relationships with others. 
To capture global information, anchor tokens are required to contain the information of all cell. This is achieved by optimizing the tokens through reconstruction of the original cell expression profiles. Specifically, we first encode the input cells into the anchor token space as:
\begin{equation}
    \bm{H} = \bm{W}_e\bm{X} + \bm{b}_e,
    \label{eq:pro_ts}
\end{equation}
where $\bm{W}_e$ and $\bm{b}_e$ are the parameters of the encoder, the $i$th row of $\bm{H}$, $\bm{h}_i$, is the embedding of cell $i$. The cell embeddings are then mapped to one of the anchor tokens, which is formulated as:
\begin{equation}
j^* = \arg\max_{j \in \{1,\dots,M\}}
\frac{\bm{h}_i \bm{u}_j^\top}{\|\bm{h}_i\|_2 \, \|\bm{u}_j\|_2}, \quad
\bm{u}_i^* = \bm{u}_{j^*},
\label{eq:map}
\end{equation}
where $\bm{u}_i^*$ is the nearest anchor token for cell $i$ in the anchor token space. We feed $\bm{u}_i^*$ to a decoder to reconstruct the raw cell expression profiles to optimize $\bm{U}$, which is denoted as:
\begin{equation}
        \bm{h}^d_i = \bm{W}_d \bm{u}_i^* + \bm{b}_d,
\end{equation}
$\bm{W}_d$ and $\bm{b}_d$ are parameters of the decoder. 
Due to the discrete, over-dispersed, and sparse nature of scRNA-seq data, directly reconstructing the expression matrix is challenging. Following prior work based on the zero-inflated negative binomial (ZINB) distribution \cite{zinb}, we estimate output mean $\theta_i$, inverse variance $\mu_i$, and dropout probability $\pi_i$ from the anchor tokens:

\begin{equation}
    \begin{aligned}
        \pi_i &= sigmoid(\bm{W}_\pi \bm{u}_i^* + \bm{b}_\pi), \\
        \theta_i &= softplus(\bm{W}_\theta \bm{u}_i^* + \bm{b}_\theta),\\
        \mu_i &= exp(\bm{W}_\mu \bm{u}_i^* + \bm{b}_\mu),
    \end{aligned}
\end{equation}
where $\bm{W}_*$ and $\bm{b}_*$ are parameters matrices. Then, the reconstruction loss function is defined as the negative log-likelihood under the ZINB model:

\begin{equation}
    \mathcal{L}_{d} = -\frac{1}{N} \sum_{i=1}^N log(ZINB(\hat{\bm{x}_i}|\pi_i, \mu_i, \theta_i)),
    \label{eq:rec_u}
\end{equation}
where $\bm{x}_i$ is the $i$-th of gene expression matrix $\hat{\bm{X}}$. To further encourage the anchor tokens to capture the semantics of the cell embeddings, we introduce a commitment loss:
\begin{equation}
    \mathcal{L}_{com}=\frac{1}{N}\sum_{i=1}^N||\bm{h}_i - \bm{u}_i^*||_2^2.
    \label{eq:com}
\end{equation}
The total loss function for anchor token learing is:
\begin{equation}
    \mathcal{L}_{a} = \mathcal{L}_{d} + \mathcal{L}_{com}.
    \label{eq:total_rec}
\end{equation}

\subsection{Bipartite Graph Attention}
 % We assume that the aggregation of a node over all cells can be approximated by interactions with the anchor tokens.
% In the conventional Transformer, the self-attention can be  
 
We propose a novel bipartite graph attention (Bi-attention) mechanism that constructs a bipartite graph between cells and anchor tokens. 
% This graph facilitates the derivation of pairwise cell similarities, achieving an approximation between bipartite graph attention and the self-attention mechanism.
% from their relationships with anchor tokens. 
% As a result, cells with similar characteristics tend to have similar anchor token embeddings, making it easier to group them into distinct clusters.
% in which each cell aggregates information from others by interacting with the anchor tokens. Since neither cell–cell nor anchor–anchor interactions are involved, the resulting attention graph naturally forms a bipartite structure rather than a fully connected graph. 
The similarity matrix in the bipartite graph is calculated by:
% The calculation for attention is defined by
\begin{equation}
\begin{aligned}
    % \bm{K} &= \bm{W}_K\bm{U}, \\
    % \bm{V} &= \bm{W}_V\bm{U}, \\
    \bm{B} &= softmax(\bm{X} \bm{W}_p (\bm{U} \bm{W}_k)^T), \\
   \bm{Z}_{out} &= \bm{B} \bm{U} \bm{W}_v,    
\end{aligned}  
\label{eq:head}
\end{equation}

where $\bm{W}_p$ is used to map cells into low-dimensional embeddings, $\bm{W}_k$ and $\bm{W}_v$ are learned parameter matrices that map the anchor tokens $\bm{U}$ into the key and value spaces, respectively, and $\bm{Z}_{out}$ is the output representation after aggregating information. Here, $\bm{B}$ is the attention matrix measuring the similarity between cells and anchor tokens, which serves as the weighted adjacency matrix of the bipartite graph. 
% Here, $f_e(\cdot)$ is an encoder that projects cells into the latent space, while $f_k(\cdot)$ and $f_v(\cdot)$ are shallow neural networks that transform meta-cells into key and value memories, respectively. 
% In this way, meta-cells serve as global memory slots shared across all cells, allowing efficient external information propagation.

% Specifically, meta-cells are treated as shared external memory, which is transformed into external memory parameters by:

% \begin{equation}
%     \bm{M}_k = f_k(E),  \bm{M}_v = f_v(E),
% \end{equation}
% where $f_k(\cdot)$ and $f_v(\cdot)$ are shallow neural network. Then, the cell embeddings are calculated by:
% \begin{equation}
% \begin{aligned}
%     \bm{A} &= softmax(\bm{Z}\bm{M}_k),\\
%     \bm{Z}_{out} &= \bm{A}\bm{M}_v,
% \end{aligned}
% \end{equation}
% where $\bm{A}$ is the attention matrix, and $\bm{Z}_{out}$ is the output of the attention mechanism.

% To further enhance the capacity of the model to capture diverse interactions, we extend the above formulation to a multi-head bipartite attention mechanism:

% \begin{equation}
% \begin{aligned}
%     \bm{Z}_{out} &= scMultiHeadEA(\bm{X}, \bm{E}) \\
%     &= \mathrm{Concat}\!\left(\hat{\bm{Z}}_{out}^1, \hat{\bm{Z}}_{out}^2, \dots, \hat{\bm{Z}}_{out}^l \right),
% \end{aligned}
% \end{equation}

To further enhance the capacity of the model to capture diverse interactions, we extend the bipartite attention in Eq.~\eqref{eq:head} to a multi-head setting, where each head constructs an independent bipartite graph between cells and anchor tokens in a distinct representation subspace:
\begin{equation}
\begin{aligned}
\bm{B}^{(i)} &= softmax(\bm{X} \bm{W}_p^{(i)} (\bm{U} \bm{W}_k^{(i)} )^\top), \\
\hat{\bm{Z}}_{out}^{(i)} &= \bm{B}^{(i)} \bm{U} \bm{W}_v^{(i)}, \\
\bm{Z}_{out} &= concat(\hat{\bm{Z}}_{out}^{(1)}, \hat{\bm{Z}}_{out}^{(2)}, \dots, \hat{\bm{Z}}_{out}^{(l)}),
\end{aligned}
\label{eq:cal_bi}
\end{equation}

where $\bm{W}_p^{(i)}$, $\bm{W}_k^{(i)}$ and $\bm{W}_v^{(i)}$ are the projection matrices of the $i$-th head, $\bm{B}^{(i)}$ denotes the corresponding attention matrix, and $l$ denotes the number of heads. Each head learns a distinct bipartite structure between cells and anchor tokens, enabling the model to capture heterogeneous and complementary relationships between cells and anchor tokens. The final embeddings can be obtained by
\begin{equation}
    \bm{Z} = \bm{Z}_{out} + \bm{X}\bm{W}_c,
    \label{eq:rep}
\end{equation}
where $\bm{Z}$ represents the learned embeddings for clustering, and $\bm{W}_c$ denotes the learnable parameters.

\subsection{scRNA-seq Clustering}
To optimize the model parameters, we define the loss function as:
\begin{equation}
    \mathcal{L} = \mathcal{L}_{s} + \mathcal{L}_c + \mathcal{L}_{a},
    \label{eq:total}
\end{equation}
where $\mathcal{L}_{s}$ denotes the self-supervised loss for learning discriminative representations, $\mathcal{L}_c$ is the DEC \cite{DEC} loss that encourages cluster-friendly embeddings (defined in Eq. \ref{eq:cluser}), and $\mathcal{L}_{a}$ is the reconstruction loss for learning anchor tokens (defined in Eq. \ref{eq:total_rec}). Here, self-supervised loss is formulated as a reconstruction objective (similar to Eq. \ref{eq:rec_u}), in which each cell is reconstructed from its learned embedding $\bm{Z}$. The BGFormer algorithm and its workflow are described in detail in Algorithm \ref{algorithmic}.

\begin{algorithm}[h]
  \caption{BGFormer-based clustering for scRNA-seq data}
  \label{alg:example}
  \begin{algorithmic}
    \STATE {\bfseries Input:} Gene expression matrix $\hat{\bm{X}} \in \mathbb{R}^{n \times d'}$, processed gene matrix ${\bm{X}} \in \mathbb{R}^{n \times d}$,  number of training epoch $\tau$.
    % number of anchor tokens $m$, number of classes $c$,
    \STATE {\bfseries Output:} Predicted cluster labels $\hat{\bm{y}}$.
    \STATE Initialize model parameters, class center $\bm{C}$ and anchor tokens $\bm{U}$.
    \FOR{$i=1$ {\bfseries to} $\tau$}

    \item Learning cell embeddings with $\bm{U}$ using Eq.~(\ref{eq:cal_bi}).

    \item Encode $\bm{X}$ to obtain representations  $\bm{H}$ using Eq. (\ref{eq:pro_ts}). \\
    
    \item Assign $\bm{H}$ to anchor tokens using Eq.~(\ref{eq:map}). \\

    \item Obtain embedding $\bm{Z}$ for clustering using Eq.~(\ref{eq:rep}). 

    % \item Calculate the reconstruction loss and the commitment loss to optimize anchor tokens by Eq. (\ref{eq:total_rec}).\\

    \item Calculate the total loss $\mathcal{L}$ by Eq. (\ref{eq:total}).

    \item Update model parameters, cluster centers $\bm{C}$ and anchor tokens $\bm{U}$ by minimizing $\mathcal{L}$.
    \ENDFOR \\
    Predict cluster labels $\hat{\bm{y}}$.
    \item return $\hat{\bm{y}}$
  \end{algorithmic}
\end{algorithm}
% Algorithm \ref{alg:example}.

% Since BGFormer models interactions between cells and anchor tokens, the attention mechanism is defined on a bipartite graph with $n \times m$ edges. As a result, the time complexity of the attention mechanism is  $\mathcal{O}(nmd)$, and the memory complexity is $\mathcal{O}(nm)$. Since $m \ll n$, the overall computational complexity of BGFormer scales linearly with respect to the number of cells, i.e., $\mathcal{O}(n)$. Compared to $\mathcal{O}(n^2)$ of standard Transformers, BGFormer is better suited for large-scale scRNA-seq clustering.

When applying scBGFormer to large-scale scRNA-seq datasets, we adopt a mini-batch training strategy. Under this setting, conventional self-attention degenerates into a local attention mechanism, as each cell can only attend to other cells within the same batch. In contrast, the proposed bipartite graph attention is well suited to large-scale datasets. The learnable anchor tokens are shared across all mini-batches and are optimized to encode global information from the entire dataset. Consequently, interactions between cells and anchor tokens can effectively approximate aggregation over all cells, even under mini-batch training. This design allows bipartite graph attention to preserve global contextual information, making scBGFormer both scalable and effective for large-scale scRNA-seq analysis.

\section{Theoretical Analysis}
When processing large-scale data, an effective solution is to employ a mini-batch training strategy. In this section, we provide a theoretical analysis to demonstrate that our proposed bipartite graph attention can effectively approximate full self-attention within each mini-batch.

% We believe that this approximation is ensured by the quality of anchor tokens and the expressive power of the multi-head attention mechanism.

\begin{theorem}
% For any $\epsilon > 0$ and $\delta > 0$, 
% there exists $k=\Theta(\frac{log(m)}{\epsilon^2})$ such that 
For any $\bm{Q}_b \in \mathbb{R}^{n' \times d}$ and $\bm{K}, \bm{V} \in \mathbb{R}^{n \times d}$, for any column vector $\bm{\omega} \in \mathbb{R}^{n}$ of matrix $\bm{V}$, there exists a low-rank matrix $\bm{\tilde{A}_b} \in \mathbb{R}^{n' \times n}$ such that
\begin{equation}
    Pr(\Vert\bm{{A}_b} \bm{\omega}^T - \bm{\tilde{A}_b} \bm{\omega}^T \Vert < \epsilon \Vert\bm{{A}_b} \bm{\omega}^T \Vert ) > 1-o(1),
\end{equation}
where $\bm{{A}}_b=softmax(\bm{Q}_b \bm{K}^T / \sqrt{d_k})$ is the attention matrix,
% where $\bm{{A}}_b$ is the attention matrix calculated by Eq.(\ref{eq:emb_weight}), 
$n'$ is the number of cells in a batch,  $\epsilon > 0$ is the error.
% , and $|| \cdot ||_F$ denotes the Frobenius norm.
\end{theorem}
We prove this result using the Johnson-Lindenstrauss (JL) Lemma \cite{JL}. Let $\bm{\tilde{A}_b} = \bm{A}_b \bm{R}^T \bm{R}$, where $\bm{R} \in \mathbb{R}^{m \times n}$ is a random matrix whose entries are drawn independently from $N(0, 1/m)$. According to the JL lemma, for any column vector $\bm{\omega} \in \mathbb{R}^{n}$ of matrix $\bm{V}$, when $m=5log(n')/(\epsilon^2-\epsilon^3)$, we have:
\begin{equation}
    Pr(||\bm{A}_b\bm{R}^T\bm{R}\bm{\omega}-\bm{A}_b\bm{\omega}|| \le \epsilon||\bm{A}_b\bm{\omega}||) > 1-o(1).
\end{equation}
Appendix \ref{sec:prove} shows more details.
Thus, a low-rank matrix can approximate the attention matrix $\bm{A}_b$ within an error of $\epsilon$.
While SVD-based low-rank approximations offer a feasible solution for approximating the attention matrix $\bm{A}_b$, the computational cost of performing SVD for each self-attention matrix can be prohibitive. To address this issue, we learn a bipartite graph attention matrix $\bm{B}$ based on anchor tokens, which approximate the full self-attention mechanism.  

% Therefore, a low-rank matrix $\bm{P}_b^{low}$ can approximate $\bm{P}_b$ using  singular value decomposition (SVD) with $\epsilon$ erro and $\mathcal{O}(mk)$ omputational complexity. However, the process of SVD decomposition in each self-attention matrix costs additional complexity. Therefore, we propose another approach for learning low-rank approximation of self-attention matrix.
% \subsection{}
% The bipartite graph attention can approximate full self-attention when t

% \begin{assumption}
% \label{as:1}
%     The anchor tokens $\bm{U}$ are said to be information-sufficient if there exists a linear decoding matrix $\bm{D}$ such that:
%     \begin{equation}
%         \lVert\bm{X} - \bm{U}\bm{D}\lVert_F \leq \varepsilon,
%     \end{equation}
%     where $\varepsilon \geq 0$ denotes the reconstruction error.
% \end{assumption}

% \begin{theorem}
%   \label{thm:bigtheorem}
%   If the anchor tokens $\bm{U}$ can linearly recover the information of all cells $\bm{X}$ up to error $\varepsilon$, then bipartite graph attention approximates full self-attention with error $O(\varepsilon)$.
% \end{theorem}
% \subsection{Anchor Token Optimization via Reconstruction}

% We provide a theoretical guarantee showing that enforcing the reconstruction loss in Eq.~(\ref{eq:rec_u}) during anchor token learning enables bipartite graph attention to approximate full self-attention.

\begin{theorem}
  \label{thm:bigtheorem}
  If the anchor tokens $\bm{U}$ can linearly recover the information of all cells $\bm{X}$, the proposed bipartite graph attention can approximate full self-attention.
\end{theorem}

The reconstruction loss in Eq.~(\ref{eq:rec_u}) guarantees that the anchor tokens effectively recover the information from all cells.
To prove Theorem \ref{thm:bigtheorem}, we reformulate the approximation of self-attention as a problem of representation approximation. Specifically, we assume that there is a linear decoding matrix $\bm{D}$ such that:
\begin{equation}
    \Vert\bm{X} - \bm{U}\bm{D}\lVert_F \leq \delta,
\end{equation}
where $\delta \geq 0$ denotes the reconstruction error. Under this assumption, the key and value matrices in Transformer can be expressed as:
\begin{equation}
    \bm{K} = (\bm{U}\bm{D})\bm{W}^K, \bm{V} = (\bm{U}\bm{D})\bm{W}^V.
\end{equation}
Following prior work \cite{soft, softmaxvibid}, we remove the softmax function in the self-attention. The representation produced by self-attention in Eq.~(\ref{eq:emb}) can then be written as:
\begin{equation}
    \begin{aligned}
        \hat{\bm{Z}} &= \bm{Q}\bm{K}^T\bm{V} \\
         &\approx \bm{Q} (\bm{U}\bm{D}\bm{W}_K)^T(\bm{U}\bm{D}\bm{W}_V) \\
         & = \bm{Q} (\bm{D}\bm{W}_K)^T\bm{U}^T\bm{U}(\bm{D}\bm{W}_V).
    \end{aligned}   
    \label{eq:the_full}
\end{equation}
Similarly, the representation learned by the bipartite graph attention in Eq.~(\ref{eq:head}) can be expressed as
\begin{equation}
    \begin{aligned}
        \bm{Z}_{out} &=  \bm{X} \bm{W}_p (\bm{U} \bm{W}_k)^T  \bm{U}\bm{W}_v \\
        & = \bm{Q} \bm{W}_k^T \bm{U}^T \bm{U} \bm{W}_v.
    \end{aligned}    
    \label{eq:the_bi}
\end{equation}

When $\bm{W}_k = \bm{D}\bm{W}_K$ and $\bm{W}_v = \bm{D}\bm{W}_V$, Eq.~(\ref{eq:the_full}) and Eq.~(\ref{eq:the_bi}) become identical in form. Therefore, the proposed bipartite graph attention is able to approximate full self-attention.

% \subsection{Stability of Multi-Head Attention}
% Next, we will discuss the stability of multi-Head attention mechanism from a linear algebraic perspective. For clarity, we ignore the softmax normalization \cite{soft}. The output of a fully attention layer can be written as:
% \begin{equation}
%     \bm{Z} = \bm{Q}\bm{M},\quad \text{where }\bm{M} = \bm{K}^T \bm{V}.
% \end{equation}

% Similarly, the multi-head bipartite graph attention is expressed as:
% \begin{equation}
%     \bm{Z}_{out} = \bm{Q} \tilde{\bm{M}}, \quad \text{where }\tilde{\bm{M}} = \bm{W}_k^T \bm{U}^T \bm{U} \bm{W}_v.
% \end{equation}

\section{Experiments}

\subsection{Datasets}
We evaluate the proposed method on several widely used single-cell RNA sequencing (scRNA-seq) datasets, including Chen \cite{Chen}, Bach \cite{Bach}, MRCA \cite{mrca}, HRCA \cite{hrca}, Fetal-Atlas \cite{fetal}, Ratmap \cite{map}, and Astrocyte \cite{Astro}. Most of these datasets contain more than 330,000 cells. 
Detailed statistical information is summarized in Table~\ref{tab:dataset}.
\begin{table}[!t]
  \caption{Summary of the scRNA-seq datasets.}
  \label{tab:dataset}
  \begin{center}
    \begin{small}
      \begin{sc}
        \begin{tabular}{crrr}
        \toprule
        Dataset     & \#Cell & \#Genes & \#Group \\ \midrule
        Chen        & 12089  & 2500    & 46      \\
        Bach        & 23184  & 1500    & 8       \\
        MRCA        & 330930 & 21255   & 27      \\
        HRCA        & 399605 & 34217   & 5       \\
        Fetal-Atlas & 433695 & 47058   & 58      \\
        Ratmap      & 504278 & 31110   & 33      \\
        Astrocyte   & 597668 & 26431   & 9       \\
        Arabidopsis & 940889 & 33341   & 9 \\   \bottomrule
        \end{tabular}
      \end{sc}
    \end{small}
  \end{center}
  \vskip -0.1in
\end{table}

\begin{table*}[!t]
  \caption{Clustering performance comparison on scRNA-seq datasets. Bold values denote the best results (\%).}
  \begin{center}
    \begin{small}
      \begin{sc}
\setlength{\tabcolsep}{4pt}
\renewcommand{\arraystretch}{1.2} 
    \begin{tabular}{c|cc|cc|cc|cc|cc|cc|cc}
\toprule
Method & \multicolumn{2}{c}{Chen} & \multicolumn{2}{c}{Bach} & \multicolumn{2}{c}{HRCA} & \multicolumn{2}{c}{MRCA} & \multicolumn{2}{c}{Fetal-Atlas} & \multicolumn{2}{c}{Ratmap} & \multicolumn{2}{c}{Astrocyte} \\
       & ACC & ARI & ACC & ARI & ACC & ARI & ACC & ARI & ACC & ARI & ACC & ARI & ACC & ARI \\
\midrule
K-means & 16.06 & 37.89 & 56.77 & 48.43 & 41.16 & 26.44 & 53.56 & 46.83 & 46.78 & 5.38 & 53.32 & 39.50 & 39.72 & 13.51 \\
Leiden & 51.56 & 37.20 & 90.58 & 87.75 & 49.05 & 8.81 & 83.51 & 78.98 & 52.49 & 34.42 & 61.16 & 45.71 & 57.38 & 30.42 \\
scICE & 64.00 & 52.81 & 88.25 & 88.35 & OOM & OOM & OOM & OOM & OOM & OOM & OOM & OOM & OOM & OOM \\
\hline
IDEC & 18.01 & 23.65 & 54.24 & 77.86 & 51.68 & 19.64 & 57.70 & 58.02 & 40.81 & 31.76 & 46.96 & 32.94 & 44.74 & 20.92 \\
scMDC & 69.39 & 68.52 & 48.09 & 83.13 & 32.98 & 5.36 & 47.97 & 62.97 & 51.79 & 40.18 & 52.38 & 31.37 & 61.52 & 46.73 \\
scDCC & 64.84 & 59.30 & 74.47 & 74.20 & 43.74 & 6.43 & 54.61 & 54.11 & 46.20 & 21.03 & 54.09 & 34.56 & 62.15 & 40.52 \\
MetaQ & 68.19 & 67.46 & 80.83 & 76.67 & 49.63 & 25.01 & 67.01 & 66.29 & 53.49 & 35.18 & 51.90 & 34.83 & 46.13 & 17.23 \\
\hline
scGNN & 51.24 & 63.98 & 87.43 & 89.11 & OOM & OOM & OOM & OOM & OOM & OOM & OOM & OOM & OOM & OOM \\
scTAG & 60.52 & 54.82 & 83.74 & 80.15 & OOM & OOM & OOM & OOM & OOM & OOM & OOM & OOM & OOM & OOM \\
CCST & 75.14 & \textbf{86.96} & 78.66 & 80.78 & OOM & OOM & OOM & OOM & OOM & OOM & OOM & OOM & OOM & OOM \\
scTPF & 70.85 & 78.99 & 85.13 & 85.20 & OOM & OOM & OOM & OOM & OOM & OOM & OOM & OOM & OOM & OOM \\
scGCL & 64.43 & 45.68 & 82.59 & 88.65 & OOM & OOM & OOM & OOM & OOM & OOM & OOM & OOM & OOM & OOM \\
\makecell{scG-\\cluster} & 73.26 & 82.40 & 73.26 & 65.94 & OOM & OOM & OOM & OOM & OOM & OOM & OOM & OOM & OOM & OOM \\
scSimGCL & 36.32 & 25.24 & 74.93 & 79.53 & 54.24 & 31.05 & 51.38 & 41.51 & 40.90 & 31.06 & 41.16 & 32.97 & 43.40 & 20.50 \\

\hline
\textbf{Ours} & \textbf{80.20} & 86.80 & \textbf{91.64} & \textbf{90.03} & \textbf{68.18} & \textbf{48.70} & \textbf{89.54} & \textbf{90.24} & \textbf{60.22} & \textbf{43.10} & \textbf{64.10} & \textbf{52.17} & \textbf{70.34} & \textbf{50.41} \\
\bottomrule
\end{tabular}

      \end{sc}
        \end{small}
  \end{center}
  \vskip -0.1in
\label{tab:clu}
\end{table*}

\subsection{Baselines}
\label{sec:baseline}
% To evaluate the performance of BGFormer, we compare it with several representative baseline methods, which can be categorized as follows:
% \begin{itemize}
%     \item  Deep-embedding methods: These methods individually encode each cells with shallow neural networks, including scDCC \cite{scDCC}, scMDC \cite{scMDC}, and MetaQ \cite{metaq}. MetaQ is designed for large-scale datasets;
%     \item GNN-based methods: These methods generate kNN graph among all cells and aggregate cell–cell relationships using graph neural networks, including scGNN \cite{scgnn}, scTAG \cite{zinb}, scTPF \cite{scTPF}, scGCL \cite{scgcl}, CCST \cite{ccst}, scG-cluster \cite{scG_cluster}, and scSimGCL \cite{scG_cluster}.
%     \item General baselines for cell clustering, including K-means, Leiden \cite{louvain} and scICE \cite{scice}.
% \end{itemize}

To evaluate the performance of BGFormer, we compare it with several representative baseline methods for single-cell clustering. For example, \emph{general clustering baselines} that commonly used in single-cell analysis, including K-means, Leiden \cite{louvain}, and scICE \cite{scice}.
\emph{Deep embedding methods} encode cells independently without explicitly modeling cell–cell interactions, including IDEC \cite{idec}, scDCC \cite{scDCC}, scMDC \cite{scMDC}, and MetaQ \cite{metaq}, where MetaQ is designed for large-scale datasets. \emph{GNN-based methods} construct a $k$-nearest neighbor (kNN) graph over all cells and apply graph neural networks to capture cell–cell relationships, including scGNN \cite{scgnn}, scTAG \cite{zinb}, CCST \cite{ccst}, scTPF \cite{scTPF}, scGCL \cite{scgcl},  scG-cluster \cite{scG_cluster},  and scSimGCL \cite{scSimGCL}. Since scGraphFormer \cite{scgraphformer} requires labeled data during training, it is unsuitable for clustering tasks. Therefore, we exclude it from our comparisons.

We evaluate clustering performance using two standard metrics:  accuracy (ACC) \cite{DEC}, and adjusted Rand index (ARI) \cite{ARI}. 
% The batch size is set to 1,000 for all datasets. 
% We use the Adam optimizer, with the learning rate selected from \{1e-3, 1e-4, 1e-5\}.
All methods are implemented using the PyTorch framework, and experiments are conducted on an NVIDIA GeForce RTX 4090 GPU with 24 GB of memory.

\subsection{Clustering Performance}
Table \ref{tab:clu} summarizes the clustering performance of BGFormer and several baseline models on scRNA-seq datasets. The best values are highlighted in bold, and “OOM” indicates an out-of-memory error. The results indicate that BGFormer almost achieves the highest ACC and ARI across all datasets. Although deep embedding methods can be easily applied to large-scale data, their performance is typically constrained by the poor quality of cell representations learned through independent encoding. GNN-based methods demonstrate superior performance on small-scale datasets. However, they are limited by the high memory cost of kNN graph construction, which can result in out-of-memory errors when applied to large-scale datasets.
BGFormer combines the advantages of two kinds of methods. It captures cell relationships by learning a bipartite graph for clustering, where pairwise cell similarities are approximated through the learned relationships between cells and anchor tokens. By avoiding direct interactions among cells, this design preserves strong scalability by enabling independent encoding of individual cells. As a result, BGFormer achieves superior clustering performance and scales effectively to large-scale scRNA-seq datasets.

% \usepackage{multirow}
% \usepackage[normalem]{ulem}
% \useunder{ ine}{ }{}

\begin{table}[]
\caption{Complexity analysis of different models.}
\setlength{\tabcolsep}{4pt} 
  \begin{center}
    \begin{small}
      \begin{sc}
\begin{tabular}{c|cc}
\toprule
            & Time Complexity & Space Complexity \\
            \midrule
% scTAG       &                 &                  \\
% \makecell{SCG- \\cluster} &                 &                  \\
scGNN       &  $\mathcal{O}(n^2d + nkd^2)$      &  $\mathcal{O}(n^2+knd)$                 \\
% scSimGCL    & $\mathcal{O}(n^2d + nkd^2 + ndc)$  &    $\mathcal{O}(n^2+knd+nc)$           \\
Transformer       &   $\mathcal{O}(n^2d+nd^2)$   &    $\mathcal{O}(n^2+nd)$   \\
Ours        & $\mathcal{O}(nmd)$  &    $\mathcal{O}(nm)$ \\
\bottomrule
\end{tabular}
      \end{sc}
        \end{small}
  \end{center}
  \vskip -0.2in
  % \vspace{-10pt}
\label{tab:complex}
\end{table}

\begin{table*}[!t]
\centering
\caption{Training and testing time per epoch (Second).}
\setlength{\tabcolsep}{7pt} 
  \begin{center}
    \begin{small}
      \begin{sc}

\setlength{\tabcolsep}{5pt} 
\begin{tabular}{c|rr|rr|rr|rr|rr|rr}
\toprule
            & \multicolumn{2}{c}{scTPF}                           & \multicolumn{2}{c}{scTAG}                           & \multicolumn{2}{c}{scG-cluster}                     & \multicolumn{2}{c}{scGNN}                        & \multicolumn{2}{c}{scSimGCL}                         & \multicolumn{2}{c}{Ours}                             \\
            & Train    & Test & Train                    & Test                     & Train                    & Test                     & Train                     & Test                 & {Train} & {Test} & {Train} & {Test} \\
            \midrule
Chen        & {3.32} & {2.33} & {1.16} & {0.21} & {6.66} & {0.98} & {11.18} & {0.65} & 1.31                     & 13.13                   & 0.81                    & 0.31                   \\
Bach        & {6.66}     & {3.25}     & {3.11} & {0.25} & {9.98} & {1.14} & {48.25} & {0.95} & 1.91                      & 5.47                   & 1.62                       & 0.47                     \\
MRCA        & -                        & -                        & -                        & -                        & -                        & -                        & -                         & -                    &  23.27                   & 166.80                   & 9.91                      & 6.89                     \\
HRCA        & -                        & -                        & -                        & -                        & -                        & -                        & -                         & -                    &    27.27                   & 49.96                    & 16.64                     & 6.54               \\
\makecell{Fetal- \\ Atlas} & -                        & -                        & -                        & -                        & -                        & -                        & -                         & -                    & 36.86                     & 441.41                   & 19.74                     & 8.08                     \\
Ratmap      & -                        & -                        & -                        & -                        & -                        & -                        & -                         & -                    & 49.37                    & 279.01                   & 23.07                     & 7.73                     \\
Astrocyte   & -                        & -                        & -                        & -                        & -                        & -                        & -                         & -                    & 60.25                     & 114.64                   & 32.53                      & 11.83   \\
\bottomrule
\end{tabular}

      \end{sc}
        \end{small}
  \end{center}
  \vskip -0.1in
\label{tab:time}
\end{table*}

% $k$ is the number of neighbors, nk 图的边
\subsection{Efficiency Analysis}
To evaluate the efficiency of BGFormer, we compare its runtime per epoch and complexity with several graph-based methods that also capture relationships among cells during clustering. Methods, such as scTAG \cite{zinb}, scTPF \cite{scTPF}, scG-cluster \cite{scTPF}, and scGNN \cite{scgnn}, require the construction of a graph before model training. scSimGCL dynamically constructs local graphs during training. 
% The results are presented in Tables \ref{tab:time} and \ref{tab:complex}. 

Table \ref{tab:complex} shows the complex analysis of these methods during embedding learning. Taking scGNN \cite{scgnn} as a representative GNN-based method, we present its time and space complexities: $\mathcal{O}(n^2d + nkd^2)$ and $\mathcal{O}(n^2 + knd)$, respectively, which account for the graph construction. 
% These complexities become prohibitive as the dataset size increases. 
Similarly, the standard Transformer model computes a similarity matrix, leading to time and space complexities of $\mathcal{O}(n^2d + nd^2)$ and $\mathcal{O}(n^2 + nd)$. These high computational and memory complexities further restrict the applicability of these models to large-scale problems. In contrast, BGFormer models interactions between cells and anchor tokens.
% , which is a bipartite graph with $n \times m$ edges. As a result, 
Its time complexity is  $\mathcal{O}(nmd)$, and the memory complexity is $\mathcal{O}(nm)$. Since $m \ll n$, the overall computational complexity of BGFormer scales linearly with the number of cells, i.e., $\mathcal{O}(n)$. Compared to these methods, BGFormer is better suited for large-scale scRNA-seq clustering.

The comparison of training and testing times per epoch, shown in Table \ref{tab:time}, further underscores the efficiency of BGFormer. The ``$\texttt{-}$" in the table indicates that these methods are unable to scale effectively with large-scale datasets due to their high computational and memory complexities. The result shows that our method significantly outperforms existing models on both small- and large-scale datasets. While scSimGCL can handle large-scale datasets, it requires longer training and testing times. Specifically, BGFormer runs 2$\times$ faster than scSimGCL during training and 10-20$\times$ faster during testing. Notably, both training and testing times for BGFormer remain stable as the number of cells increases. 
These findings indicate that BGFormer is an efficient solution for large-scale scRNA-seq clustering.

% We analyze the efficiency of BGFormer and existing methods on datasets. Since both BGFormer and GNN-based methods learn cell embeddings by modeling relationships among cells, we focus the comparison on these approaches. Table~\ref{tab:time} presents the per-epoch runtime during training. On large-scale datasets, many GNN-based methods fail due to memory constraints, whereas BGFormer remains stable. On small-scale datasets, BGFormer achieves the shortest runtime. On large-scale datasets, it runs 10–20$\times$ faster than scSimGCL, which dynamically constructs local graphs during training.

% \begin{figure}[!t]
%     \centering
%     \includegraphics[width=1\columnwidth]{figs/vis/time_scatter.pdf}
%     \caption{Comparison of runtime across datasets of increasing size. Dashed lines indicate methods that fail due to out-of-memory errors on large datasets.}
%     \label{fig:time}
% \end{figure}

% \subsection{Class-Level Attention Visualization}
\subsection{Visualization}

\begin{figure}[!t]
    \centering
    \includegraphics[width=1\columnwidth]{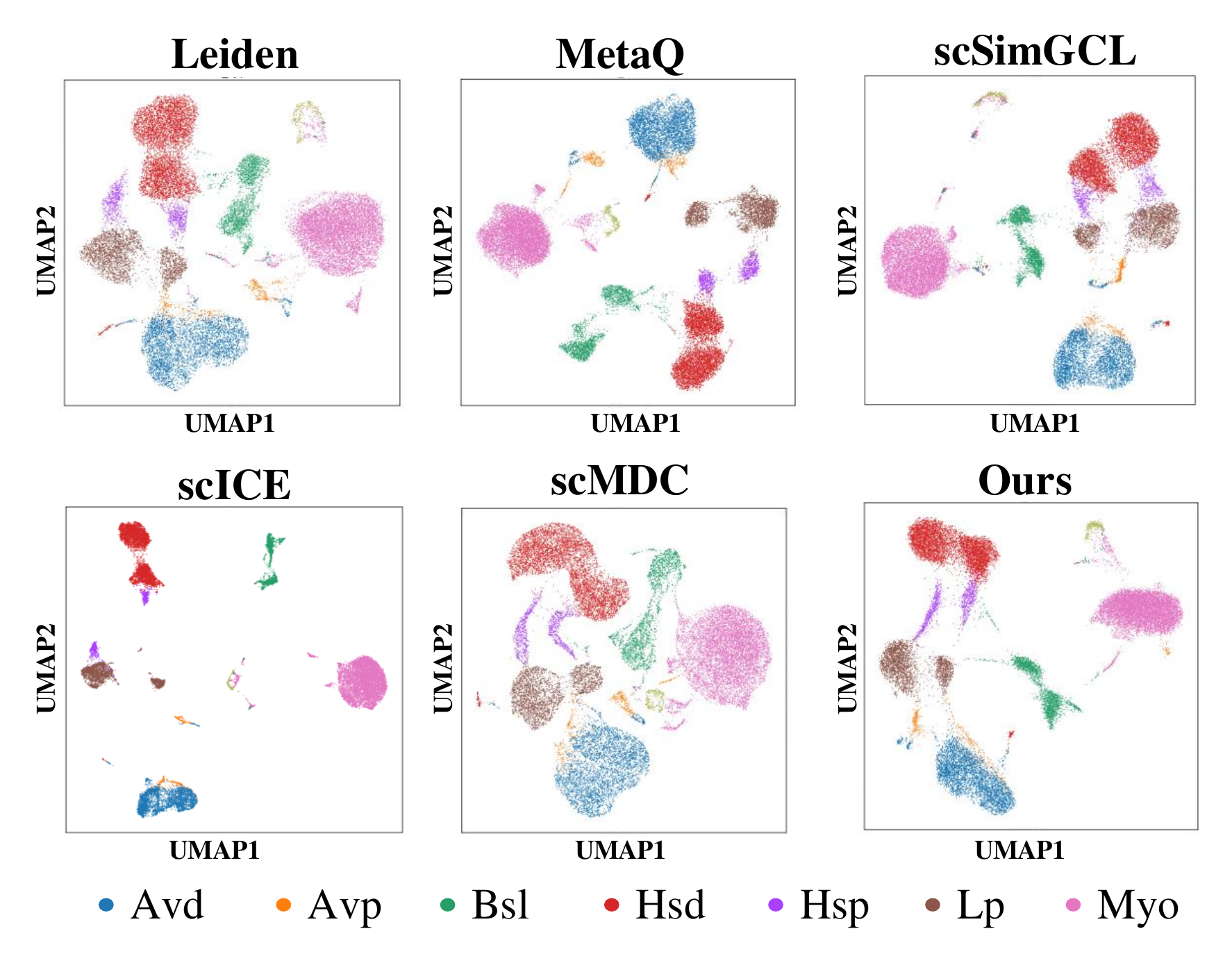}
    \caption{UMAP visualizations of cell embeddings, with colors indicating cell types.}
    \label{fig:vis}
\end{figure}

To evaluate the discrimination of the learned representations, we perform a UMAP visualization \cite{Umap} on the Bach dataset. In the visualization, cells with the same ground-truth labels are assigned the same color. We compare our model with some representative models from three kinds of methods.
% discussed in Section \ref{sec:baseline}. 
As shown in Fig. \ref{fig:vis}, our model produces well-separated and coherent clusters, demonstrating the effectiveness of the learned representations in distinguishing different cell types. 
In contrast, scICE demonstrates notable misclassification, while other methods suffer from  cellular admixture and cluster overlap.

% To better understand how the multi-head attention mechanism captures relationships between cells and anchor tokens, we visualize the class-level attention weight distributions across different attention heads. Specifically, attention matrices are grouped according to the ground-truth class labels, and the mean attention score is computed for each class. The resulting class-level attention heatmaps are shown in Figure~\ref{fig:attn}. As illustrated in the figure, different classes exhibit clearly distinct attention patterns toward anchor tokens, indicating that anchor tokens capture class-discriminative global information. Moreover, attention patterns within the same class vary substantially across different heads, suggesting that multi-head attention models class-specific relationships from complementary perspectives.

% \begin{figure}[!t]
%   % \vskip 0.2in
%   \begin{center}
%     \centerline{\includegraphics[width=1\columnwidth]{figs/attn/attention.pdf}}
%     \caption{
%       Class-wise attention heatmaps of different attention heads.
%     }
%     \label{fig:attn}
%   \end{center}
% \end{figure}

\begin{figure*}[!t]
    \centering
    \includegraphics[width=1\textwidth]{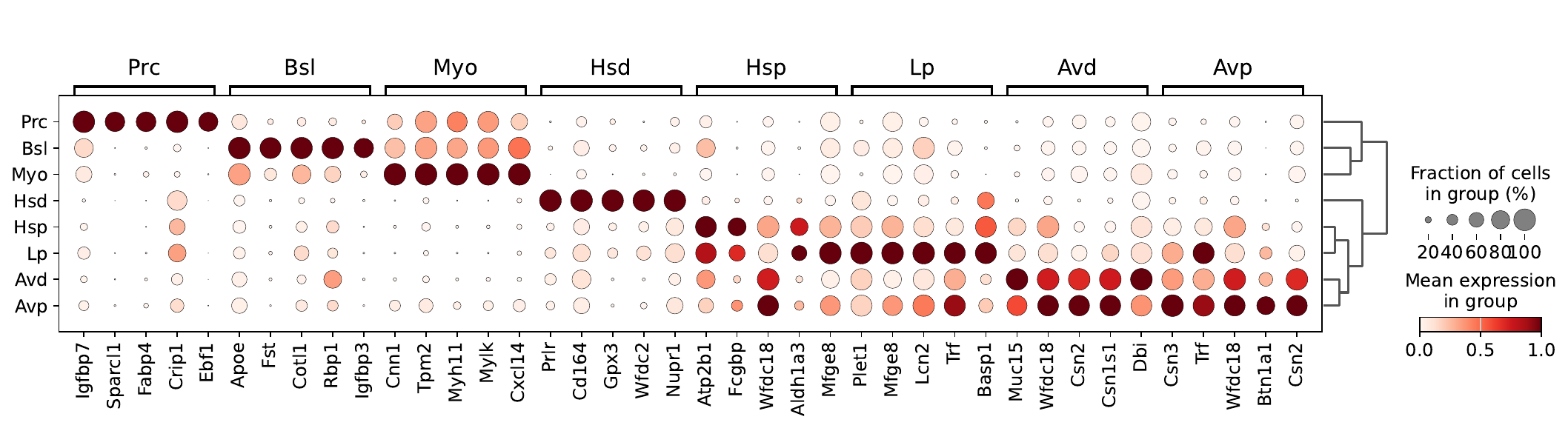}
    \caption{Dot plots showing the expression of genes on the Bach dataset.}
    \label{fig:dotplot}
\end{figure*}

% \begin{table*}[!t]
% \centering
% \caption{Ablation study on clustering (\%).}
% \setlength{\tabcolsep}{7pt} 
%   \begin{center}
%     \begin{small}
%       \begin{sc}
% \begin{tabular}{ccccccccccc}
% \toprule
% Dataset & \multicolumn{3}{c}{ACC} & \multicolumn{3}{c}{NMI} & \multicolumn{3}{c}{ARI} \\
% % \cmidrule(lr){2-4} \cmidrule(lr){5-7} \cmidrule(lr){8-10}
% & w/o $\mathcal{L}_{a}$ & w/o $\mathcal{L}_{s}$ & Full & w/o $\mathcal{L}_{a}$ & w/o $\mathcal{L}_{s}$ & Full & w/o $\mathcal{L}_{a}$ & w/o $\mathcal{L}_{s}$ & Full \\
% \midrule
% Chen     & 64.86 & 60.65 & \textbf{80.20} & 73.48 & 72.91 & \textbf{81.56} & 55.18 & 50.39 & \textbf{86.80} \\
% Bach     & 83.32 & 82.91 & \textbf{91.64} & 79.48 & 80.58 & \textbf{86.63} & 80.92 & 83.62 & \textbf{90.03} \\
% HRCA     & 63.94 & 57.00 & \textbf{68.18} & 38.39 & 36.33 & \textbf{42.50} & 43.39 & 35.17 & \textbf{48.70} \\
% MRCA     & 81.94 & 73.48 & \textbf{89.54} & 83.65 & 83.98 & \textbf{89.49} & 74.70 & 88.61 & \textbf{90.24} \\
% Fetal-Atlas & 52.54 & 46.99 & \textbf{60.22} & 64.54 & 61.04 & \textbf{66.50} & 32.41 & 28.98 & \textbf{43.10} \\
% Ratmap   & 59.77 & 56.00 & \textbf{64.10} & 63.09 & 59.80 & \textbf{66.16} & 44.89 & 39.10 & \textbf{52.17} \\
% Astrocyte & 62.44 & 32.59 & \textbf{70.34} & 43.52 & 41.65 & \textbf{55.23} & 38.72 & 12.72 & \textbf{50.41} \\
% \bottomrule
% \end{tabular}
%       \end{sc}
%         \end{small}
%   \end{center}
%   \vskip -0.1in
% \label{tab:abl}
% \end{table*}

% \subsection{Analysis of Cluster-wise Marker Gene Analysis}
Fig. \ref{fig:dotplot} presents a dot plot of the top genes ranked by mean expression across across the predicted clusters produced by BGFormer on the Bach dataset, where dot size and color denote the fraction of expressing cells and the mean expression level, respectively. The varying dot sizes between clusters indicate that BGFormer has successfully distinguished the cell populations, highlighting that it captures the heterogeneity between clusters. In addition, both Alveolar differentiated cells (Avd) and Alveolar progenitor cells (Avp) exhibit high expression of protein-coding genes (e.g., Csn2, Csn1s1 and Wfdc18),
% and are composed of cells from gestation and lactation, 
supporting the hypothesis that these are secretory alveolar cells.
% the high expression of csn2 in both Alveolar differentiated cells (Avd) and Alveolar progenitor cells (Avp) suggests that this gene may play an important role in the development and differentiation of alveolar cells. 
% Clusters enriched for Igfbp3, Igfbp7, and Col1a1 exhibit signatures associated with extracellular matrix organization and stromal cell functions. In addition, immune-related genes such as Cd164, Ctsh, and Lcn2 display cluster-specific activation patterns, suggesting functional diversity among immune or immune-adjacent cell states.

\begin{figure}[ht]
  % \vskip 0.2in
  \begin{center}
    \centerline{\includegraphics[width=1\columnwidth]{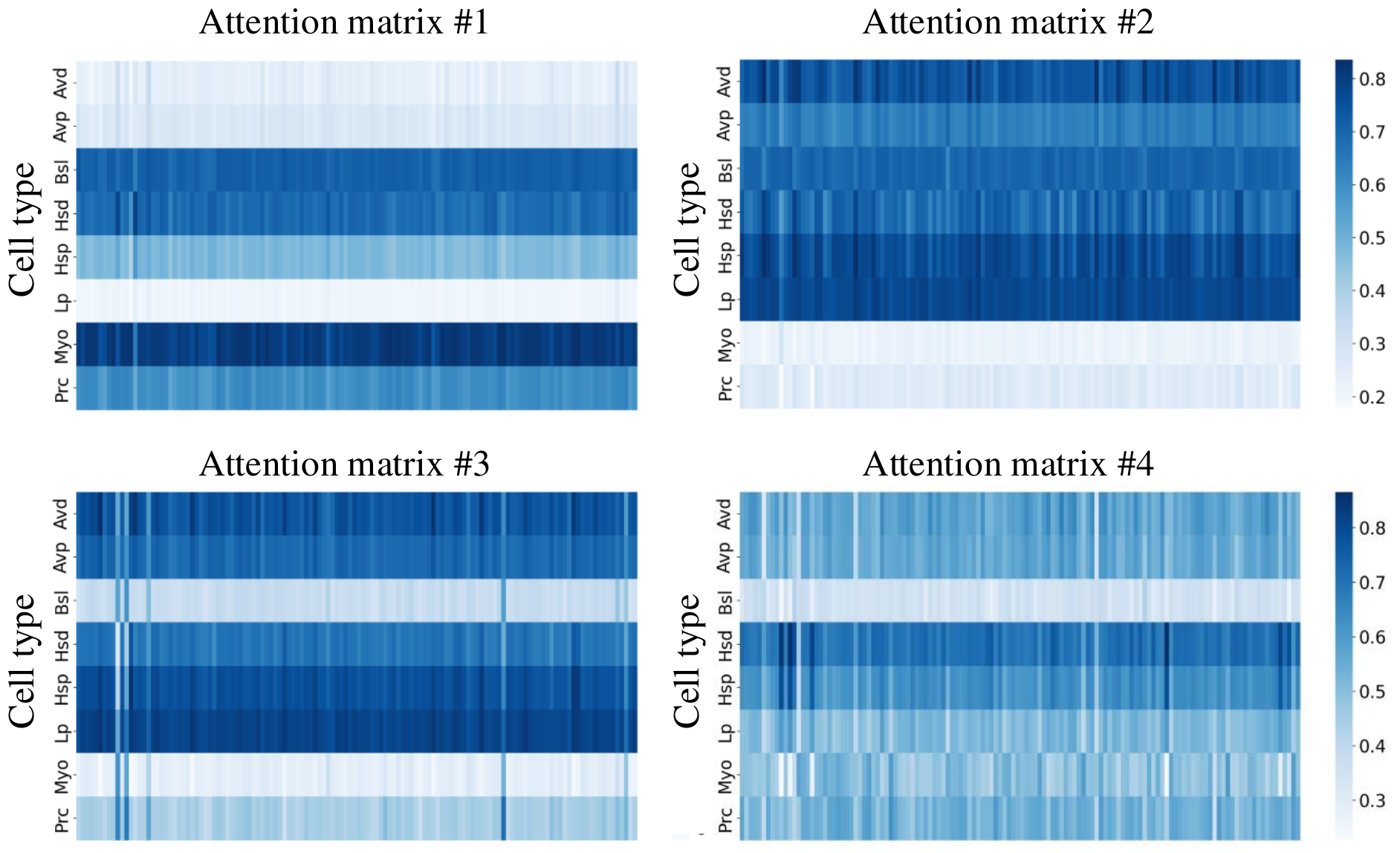}}
    \caption{
      Class-wise attention heatmaps of different attention heads.
    }
    \label{fig:attn}
  \end{center}
\end{figure}

To better understand how the multi-head attention mechanism captures relationships between cells and anchor tokens, we visualize the class-level attention weight distributions across different attention heads. Specifically, attention matrices are grouped according to the ground-truth class labels, and the mean attention score is computed for each class. The resulting class-level attention heatmaps are shown in Figure~\ref{fig:attn}. As illustrated in the figure, different classes exhibit clearly distinct attention patterns toward anchor tokens, indicating that anchor tokens capture class-discriminative global information. Moreover, attention patterns within the same class vary substantially across different heads, suggesting that multi-head attention models class-specific relationships from complementary perspectives.

\subsection{Ablation study}
To analyze the contribution of each loss component, we compare BGFormer with two variants across all datasets, where each variant omits one part of the total loss during training. Table \ref{tab:abl} presents the single-cell clustering results, with `w/o $\mathcal{L}_{a}$' indicating the exclusion of Eq. (\ref{eq:total_rec}) and `w/o $\mathcal{L}_{s}$' indicating the removal of self-supervised loss function. The results show that constraining the anchor tokens to encode the key information of all cells enables effective aggregation and propagation of global information, thereby improving clustering performance. Additionally, improving the quality of cell embeddings via self-supervised learning is critical for achieving accurate clustering. Overall, these findings demonstrate that all components of BGFormer are both necessary and effective.

\begin{table}[!t]
\centering
\caption{Ablation study on clustering (\%).}
\setlength{\tabcolsep}{7pt} 
  \begin{center}
    \begin{small}
      \begin{sc}
      \setlength{\tabcolsep}{2pt}
      \renewcommand{\arraystretch}{1.2} 
\begin{tabular}{c|ccc|ccc}
\toprule
Dataset & \multicolumn{3}{c}{ACC} & \multicolumn{3}{c}{ARI} \\
& w/o $\mathcal{L}_{a}$ & w/o $\mathcal{L}_{s}$ & Full & w/o $\mathcal{L}_{a}$ & w/o $\mathcal{L}_{s}$ & Full \\
\midrule
Chen     & 64.86 & 60.65 & \textbf{80.20} & 55.18 & 50.39 & \textbf{86.80} \\
Bach     & 83.32 & 82.91 & \textbf{91.64} & 80.92 & 83.62 & \textbf{90.03} \\
HRCA     & 63.94 & 57.00 & \textbf{68.18} & 43.39 & 35.17 & \textbf{48.70} \\
MRCA     & 81.94 & 73.48 & \textbf{89.54} & 74.70 & 88.61 & \textbf{90.24} \\
\makecell{Fetal-\\Atlas} & 52.54 & 46.99 & \textbf{60.22} & 32.41 & 28.98 & \textbf{43.10} \\
Ratmap   & 59.77 & 56.00 & \textbf{64.10} & 44.89 & 39.10 & \textbf{52.17} \\
Astrocyte & 62.44 & 32.59 & \textbf{70.34} & 38.72 & 12.72 & \textbf{50.41} \\
\bottomrule
\end{tabular}
      \end{sc}
        \end{small}
  \end{center}
  \vskip -0.1in
\label{tab:abl}
\end{table}

\subsection{Parameter Analysis}
\begin{figure}
    \centering
    \includegraphics[width=\columnwidth]{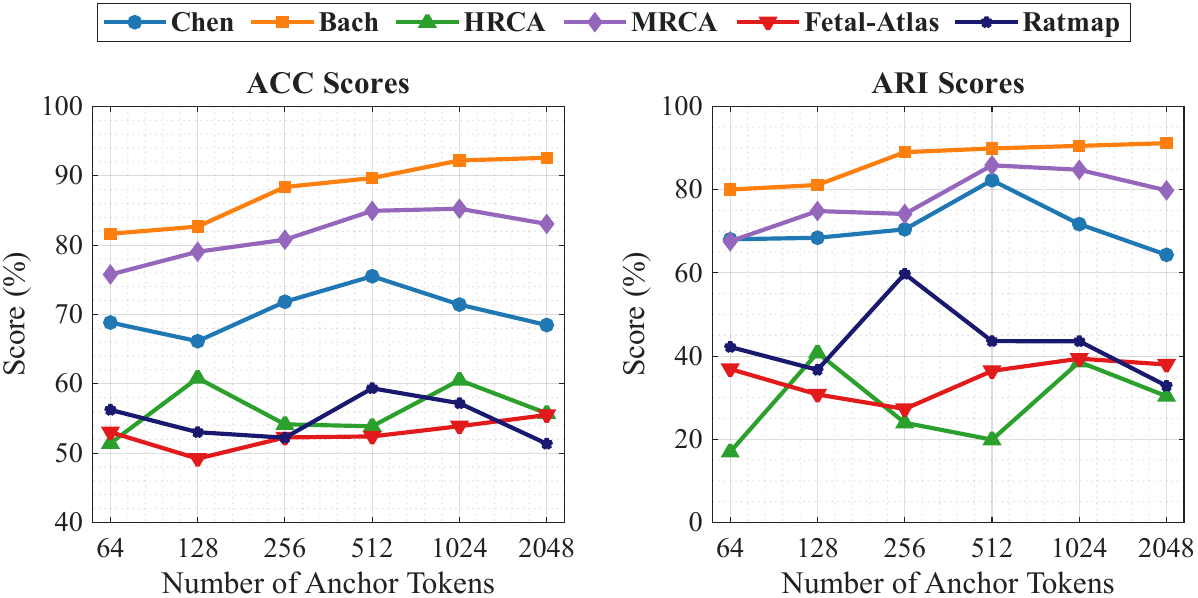}
    \caption{Analysis of the number of anchor tokens.}
    \label{fig:number_tokens}
\end{figure}

% \subsubsection{Impact of the number of anchor tokens.}
We investigate the effect of varying the number of anchor tokens, denoted as $m \in \{64,128,256,512,1024,2048\}$, on model performance.
% We investigate the impact of the number of anchor tokens on model performance by varying the anchor token size $m$ in $\{64,128,256,512,1024,2024\}$. 
As shown in Fig. \ref{fig:number_tokens}, the number of anchor tokens significantly influences clustering performance. When the number of anchor tokens is small, the anchors fail to adequately represent the complexity and heterogeneity of large scRNA-seq datasets, resulting in suboptimal clustering performance. As the number of anchor tokens increases, performance steadily improves, indicating that richer anchor representations provide more informative global references for cell embeddings. However, when the number of anchor tokens becomes too large, redundant information is introduced, which can impair model performance.

% \begin{figure}
%     \centering
%     % \includegraphics[width=\columnwidth]{figs/par_ana/head.pdf}
%     \caption{Analysis of the number of heads.}
%     \label{fig:placeholder}
% \end{figure}

\section{Conclusion}
% We propose a scalable and effective bipartite graph transformer-based clustering model (BGFormer) for single-cell RNA-seq data. 
% BGFormer approximates full self-attention with a computational complexity that scales linearly, rather than quadratically, with the number of cells. These anchor tokens are optimized to capture global cellular information, providing comprehensive representation of the entire dataset within each mini-batch. A bipartite graph attention mechanism is proposed to propagates this information to individual cells, enabling the learning of discriminative cell embeddings for clustering. 
% Experiments demonstrate that BGFormer achieves strong clustering performance with high efficiency.

We propose a scalable and effective bipartite graph transformer-based clustering model (BGFormer) for single-cell RNA-seq data. Unlike existing methods that compute similarities between all pairs of cells, BGFormer focuses on capturing the relationships between cells and a small set of learnable anchor tokens. Because the number of anchor tokens is much smaller than the number of cells, the computational complexity of BGFormer scales linearly, rather than quadratically, with the number of cells. We optimize the anchor tokens to capture global cellular information. A bipartite graph attention mechanism is proposed to learn a similarity matrix that reflects cell groupings and enhances the separation between clusters. Experiments demonstrate that BGFormer achieves superior clustering performance with high efficiency.

% \section*{Acknowledgments}
% This should be a simple paragraph before the References to thank those individuals and institutions who have supported your work on this article.



\bibliography{ref}

@article{scrnadata,
  title={scRNA-seq data analysis method to improve analysis performance},
  author={Lu, Junru and Sheng, Yuqi and Qian, Weiheng and Pan, Min and Zhao, Xiangwei and Ge, Qinyu},
  journal={IET nanobiotechnology},
  volume={17},
  number={3},
  pages={246--256},
  year={2023},
}

@inproceedings{zinb,
  title={Zinb-based graph embedding autoencoder for single-cell rna-seq interpretations},
  author={Yu, Zhuohan and Lu, Yifu and Wang, Yunhe and Tang, Fan and Wong, Ka-Chun and Li, Xiangtao},
  booktitle={Proceedings of the AAAI conference on artificial intelligence},
  volume={36},
  number={4},
  pages={4671--4679},
  year={2022}
}

@article{louvain,
  title={From Louvain to Leiden: guaranteeing well-connected communities},
  author={Traag, Vincent A and Waltman, Ludo and Van Eck, Nees Jan},
  journal={Scientific reports},
  volume={9},
  number={1},
  pages={1--12},
  year={2019},
  publisher={Nature Publishing Group}
}

@article{scgraphformer,
  title={scGraphformer: unveiling cellular heterogeneity and interactions in scRNA-seq data using a scalable graph transformer network},
  author={Fan, Xingyu and Liu, Jiacheng and Yang, Yaodong and Gu, Chunbin and Han, Yuqiang and Wu, Bian and Jiang, Yirong and Chen, Guangyong and Heng, Pheng-Ann},
  journal={Communications Biology},
  volume={7},
  number={1},
  pages={1463},
  year={2024},
  publisher={Nature Publishing Group UK London}
}

@article{scice,
  title={scICE: enhancing clustering reliability and efficiency of scRNA-seq data with multi-cluster label consistency evaluation},
  author={Kim, Hyun and Park, Issac and Park, Jong-Eun and Kim, Jong Kyoung and Seo, Minseok and Kim, Jae Kyoung},
  journal={Nature Communications},
  volume={16},
  number={1},
  pages={6031},
  year={2025},
  publisher={Nature Publishing Group UK London}
}

@article{scgnn,
  title={scGNN is a novel graph neural network framework for single-cell RNA-Seq analyses},
  author={Wang, Juexin and Ma, Anjun and Chang, Yuzhou and Gong, Jianting and Jiang, Yuexu and Qi, Ren and Wang, Cankun and Fu, Hongjun and Ma, Qin and Xu, Dong},
  journal={Nature communications},
  volume={12},
  number={1},
  pages={1882},
  year={2021},
  publisher={Nature Publishing Group UK London}
}

@article{Chen,
  title={Single-cell RNA-seq reveals hypothalamic cell diversity},
  author={Chen, Renchao and Wu, Xiaoji and Jiang, Lan and Zhang, Yi},
  journal={Cell reports},
  volume={18},
  number={13},
  pages={3227--3241},
  year={2017},
  publisher={Elsevier}
}

@article{Bach,
  title={Differentiation dynamics of mammary epithelial cells revealed by single-cell RNA sequencing},
  author={Bach, Karsten and Pensa, Sara and Grzelak, Marta and Hadfield, James and Adams, David J and Marioni, John C and Khaled, Walid T},
  journal={Nature communications},
  volume={8},
  number={1},
  pages={1--11},
  year={2017},
  publisher={Nature Publishing Group}
}

@article{mrca,
  title={Comprehensive single-cell atlas of the mouse retina},
  author={Li, Jin and Choi, Jongsu and Cheng, Xuesen and Ma, Justin and Pema, Shahil and Sanes, Joshua R and Mardon, Graeme and Frankfort, Benjamin J and Tran, Nicholas M and Li, Yumei and others},
  journal={Iscience},
  volume={27},
  number={6},
  year={2024},
  publisher={Elsevier}
}

@article{hrca,
  title={Integrated multi-omics single cell atlas of the human retina},
  author={Li, Jin and Wang, Jun and Ibarra, Ignacio L and Cheng, Xuesen and Luecken, Malte D and Lu, Jiaxiong and Monavarfeshani, Aboozar and Yan, Wenjun and Zheng, Yiqiao and Zuo, Zhen and others},
  journal={Research Square},
  pages={rs--3},
  year={2023}
}

@article{fetal,
  title={A human cell atlas of fetal gene expression},
  author={Cao, Junyue and O’day, Diana R and Pliner, Hannah A and Kingsley, Paul D and Deng, Mei and Daza, Riza M and Zager, Michael A and Aldinger, Kimberly A and Blecher-Gonen, Ronnie and Zhang, Fan and others},
  journal={Science},
  volume={370},
  number={6518},
  pages={eaba7721},
  year={2020},
  publisher={American Association for the Advancement of Science}
}

@article{map,
  title={Transcriptional profile of the rat cardiovascular system at single-cell resolution},
  author={Arduini, Alessandro and Fleming, Stephen J and Xiao, Ling and Hall, Amelia W and Akkad, Amer-Denis and Chaffin, Mark D and Bendinelli, Kayla J and Tucker, Nathan R and Papangeli, Irinna and Mantineo, Helene and others},
  journal={Cell Reports},
  volume={44},
  number={1},
  year={2025},
  publisher={Elsevier}
}

@article{Astro,
  title={Astrocyte regional specialization is shaped by postnatal development},
  author={Schroeder, Margaret E and McCormack, Dana M and Metzner, Lukas R and Kang, Jinyoung and Li, Katelyn X and Yu, Eunah and Melamed, Lisa and Levandowski, Kirsten M and Zaniewski, Heather and Zhang, Qiangge and others},
  journal={bioRxiv},
  pages={2024--10},
  year={2025}
}

@inproceedings{acc,
  title={Unsupervised deep embedding for clustering analysis},
  author={ Xie, Junyuan  and  Girshick, Ross Brook  and  Farhadi, Ali },
  booktitle={International Conference on Machine Learning},
  year={2016},
}

@article{ARI,
  title={Comparing partitions},
  author={ Hubert, Lawrence  and  Arabie, Phipps },
  journal={J. Classif},
  volume={2},
  number={1},
  pages={193-218},
  year={1985},
}

@article{scMDC,
  title={Clustering of single-cell multi-omics data with a multimodal deep learning method},
  author={Lin, Xiang and Tian, Tian and Wei, Zhi and Hakonarson, Hakon},
  journal={Nature communications},
  volume={13},
  number={1},
  pages={7705},
  year={2022},
  publisher={Nature Publishing Group UK London}
}

@article{scDCC,
  title={Model-based deep embedding for constrained clustering analysis of single cell RNA-seq data},
  author={Tian, Tian and Zhang, Jie and Lin, Xiang and Wei, Zhi and Hakonarson, Hakon},
  journal={Nature communications},
  volume={12},
  number={1},
  pages={1873},
  year={2021},
  publisher={Nature Publishing Group UK London}
}

@inproceedings{scTPF,
  title={Exploring the interaction between local and global latent configurations for clustering single-cell rna-seq: a unified perspective},
  author={Mrabah, Nairouz and Amar, Mohamed Mahmoud and Bouguessa, Mohamed and Diallo, Abdoulaye Banire},
  booktitle={Proceedings of the AAAI Conference on Artificial Intelligence},
  volume={37},
  number={8},
  pages={9235--9242},
  year={2023}
}

@article{scgcl,
  title={scGCL: an imputation method for scRNA-seq data based on graph contrastive learning},
  author={Xiong, Zehao and Luo, Jiawei and Shi, Wanwan and Liu, Ying and Xu, Zhongyuan and Wang, Bo},
  journal={Bioinformatics},
  volume={39},
  number={3},
  pages={btad098},
  year={2023},
  publisher={Oxford University Press}
}

@article{ccst,
  title={Cell clustering for spatial transcriptomics data with graph neural networks},
  author={Li, Jiachen and Chen, Siheng and Pan, Xiaoyong and Yuan, Ye and Shen, Hong-Bin},
  journal={Nature Computational Science},
  volume={2},
  number={6},
  pages={399--408},
  year={2022},
  publisher={Nature Publishing Group US New York}
}

@article{scG_cluster,
  title={Deep learning powered single-cell clustering framework with enhanced accuracy and stability},
  author={Zhang, Yi and Feng, Xi and Wang, Yin and Shi, Kai},
  journal={Scientific Reports},
  volume={15},
  number={1},
  pages={4107},
  year={2025},
  publisher={Nature Publishing Group UK London}
}

@article{scSimGCL,
  title={Graph contrastive learning as a versatile foundation for advanced scRNA-seq data analysis},
  author={Zhang, Zhenhao and Liu, Yuxi and Xiao, Meichen and Wang, Kun and Huang, Yu and Bian, Jiang and Yang, Ruolin and Li, Fuyi},
  journal={Briefings in Bioinformatics},
  volume={25},
  number={6},
  pages={bbae558},
  year={2024},
  publisher={Oxford University Press}
}

@article{metaq,
  title={MetaQ: fast, scalable and accurate metacell inference via single-cell quantization},
  author={Li, Yunfan and Li, Hancong and Lin, Yijie and Zhang, Dan and Peng, Dezhong and Liu, Xiting and Xie, Jie and Hu, Peng and Chen, Lu and Luo, Han and others},
  journal={Nature Communications},
  volume={16},
  number={1},
  pages={1205},
  year={2025},
  publisher={Nature Publishing Group UK London}
}

@inproceedings{idec,
  title={Improved deep embedded clustering with local structure preservation.},
  author={Guo, Xifeng and Gao, Long and Liu, Xinwang and Yin, Jianping},
  booktitle={Ijcai},
  volume={17},
  pages={1753--1759},
  year={2017}
}

@inproceedings{softmaxvibid,
  title={Vibid: Linear vision transformer with bidirectional normalization},
  author={Song, Jeonggeun and Lee, Heung-Chang},
  booktitle={Uncertainty in Artificial Intelligence},
  pages={1996--2005},
  year={2023},
  organization={PMLR}
}

@article{soft,
title = {Simplified Transformer},
journal = {Neurocomputing},
volume = {647},
pages = {130497},
year = {2025},
author = {Lei Xu and Haiying Luo and Zhang Yi},
}

@inproceedings{DEC,
  title={Unsupervised deep embedding for clustering analysis},
  author={Xie, Junyuan and Girshick, Ross and Farhadi, Ali},
  booktitle={International conference on machine learning},
  pages={478--487},
  year={2016},
  organization={PMLR}
}

@article{umap,
  title={Umap: Uniform manifold approximation and projection for dimension reduction},
  author={McInnes, Leland and Healy, John and Melville, James},
  journal={arXiv preprint arXiv:1802.03426},
  year={2018}
}

@inproceedings{kmeans,
  title={Some methods of classification and analysis of multivariate observations},
  author={McQueen, James B},
  booktitle={Proc. of 5th Berkeley Symposium on Math. Stat. and Prob.},
  pages={281--297},
  year={1967}
}

@inproceedings{gcn,
  author    = {Thomas N. Kipf and
               Max Welling},
  title     = {Semi-Supervised Classification with Graph Convolutional Networks},
  booktitle = {Proceedings of the 5th International Conference on Learning Representations},
  year      = {2017}
}

@article{scgae,
  title={A topology-preserving dimensionality reduction method for single-cell RNA-seq data using graph autoencoder},
  author={Luo, Zixiang and Xu, Chenyu and Zhang, Zhen and Jin, Wenfei},
  journal={Scientific reports},
  volume={11},
  number={1},
  pages={20028},
  year={2021},
  publisher={Nature Publishing Group UK London}
}

@article{scvae,
  title={scVAE: variational auto-encoders for single-cell gene expression data},
  author={Gr{\o}nbech, Christopher Heje and Vording, Maximillian Fornitz and Timshel, Pascal N and S{\o}nderby, Casper Kaae and Pers, Tune H and Winther, Ole},
  journal={Bioinformatics},
  volume={36},
  number={16},
  pages={4415--4422},
  year={2020},
  publisher={Oxford University Press}
}

@article{linformer,
  title={Linformer: Self-attention with linear complexity},
  author={Wang, Sinong and Li, Belinda Z and Khabsa, Madian and Fang, Han and Ma, Hao},
  journal={arXiv preprint arXiv:2006.04768},
  year={2020}
}

@article{nodeformer,
  title={Nodeformer: A scalable graph structure learning transformer for node classification},
  author={Wu, Qitian and Zhao, Wentao and Li, Zenan and Wipf, David P and Yan, Junchi},
  journal={Advances in Neural Information Processing Systems},
  volume={35},
  pages={27387--27401},
  year={2022}
}

@article{JL,
  title={Extensions of lipschitz maps into a hilbert space},
  author={Lindenstrauss, W Johnson J and Johnson, J},
  journal={Contemp. Math},
  volume={26},
  number={189-206},
  pages={2},
  year={1984}
}

@article{massively,
  title={Massively parallel digital transcriptional profiling of single cells},
  author={Zheng, Grace XY and Terry, Jessica M and Belgrader, Phillip and Ryvkin, Paul and Bent, Zachary W and Wilson, Ryan and Ziraldo, Solongo B and Wheeler, Tobias D and McDermott, Geoff P and Zhu, Junjie and others},
  journal={Nature communications},
  volume={8},
  number={1},
  pages={14049},
  year={2017},
  publisher={Nature Publishing Group UK London}
}

@article{transformers_cell1,
  title={Transformers in single-cell omics: a review and new perspectives},
  author={Sza{\l}ata, Artur and Hrovatin, Karin and Becker, S{\"o}ren and Tejada-Lapuerta, Alejandro and Cui, Haotian and Wang, Bo and Theis, Fabian J},
  journal={Nature methods},
  volume={21},
  number={8},
  pages={1430--1443},
  year={2024},
  publisher={Nature Publishing Group US New York}
}

@misc{transformers_cell2,
  title={Transformer for one stop interpretable cell type annotation. Nat. Commun. 14, 223},
  author={Chen, J and Xu, H and Tao, W and Chen, Z and Zhao, Y and Han, JDJ},
  year={2023}
}
\bibliographystyle{IEEEtran}

\vfill

\end{document}